\pdfoutput=1

\documentclass[11pt]{article}

\usepackage[]{acl2023}

\usepackage{times}
\usepackage{latexsym}

\usepackage[hang,flushmargin]{footmisc} 

\usepackage[T1]{fontenc}

\usepackage[utf8]{inputenc}

\usepackage{microtype}

\usepackage{inconsolata}

\title{AlbMoRe: A Corpus of Movie Reviews for Sentiment Analysis in Albanian}
\author{Erion \c{C}ano \\
  Digital Philology \\
  Data Mining and Machine Learning \\
  University of Vienna, Austria \\
  \texttt{erion.cano@univie.ac.at} \\
  }

\begin{document}

\maketitle

\begin{abstract}
Lack of available resources such as text corpora for low-resource languages seriously hinders research on natural language processing and computational linguistics. This paper presents AlbMoRe, a corpus of 800 sentiment annotated movie reviews in Albanian. Each text is labeled as \emph{positive} or \emph{negative} and can be used for sentiment analysis research. Preliminary results based on traditional machine learning classifiers trained with the AlbMoRe samples are also reported. They can serve as comparison baselines for future research experiments.   
\end{abstract}

\section{Introduction}

The growth of data-driven artificial intelligence solutions for language processing tasks has motivated the creation of big text corpora \cite{boulton:hal-01854664,10.3389/feduc.2019.00007}. These corpora usually contain a text part in a natural language and are often labeled with some extra information (e.g., a category) added by humans. As a matter of fact, most of the available research corpora have been developed for English language.

The resources like text corpora, processing software and pre-trained models created for ``smaller'' languages, also known as low-resource or underrepresented languages, are scarce. This creates difficulties, hinders progress and limits the obtained performance in modelling and automating natural language processing and computational linguistics tasks for those languages.  

This paper presents AlbMoRe, a corpus of 800 movie reviews in Albanian language, created with the goal to foster sentiment analysis research.\footnote{Download from: \url{http://hdl.handle.net/11234/1-5165}} The reviews were collected from IMDb\footnote{\url{https://www.imdb.com}} which is the most popular internet platform with information related to movies and user reviews about them. The reviews in AlbMoRe belong to 67 movies of different genres.
A set of sentiment analysis experiments were run on the corpus, assessing the classification accuracy of a few traditional machine learning models.\footnote{Code at \url{https://github.com/erionc/AlbMoRe}} The respective results should serve as comparison baselines for further experiments involving more advanced models in the future.

\section{Related Work} %

Sentiment analysis of texts is about using labeled, unlabeled, or partly labeled corpora for training intelligent models that are later used to automatically analyze the emotional polarity of text fragments. The text units that are analyzed can be short messages up to a sentence long, reviews about products up to a paragraph long or even longer units spanning up to an entire document. 

The simplest task of the trained intelligent models is to recognize the sentiment polarity of the text, which is to sort out the \emph{positive} units from the \emph{negative} ones. In reality, we often encounter \emph{neutral} texts as well. The task becomes harder when we also need to assess the degree of positivity or negativity, or when we need to know about more specific emotional states such as \emph{enjoyment}, \emph{anger}, \emph{disgust}, \emph{sadness}, \emph{fear} and \emph{surprise}. 

Sentiment analysis has been traditionally conceived as a binary or multi-class classification task, using the vector space model and the Tf-Idf weighting scheme to represent the texts \cite{Ramos1999,10.5555/1796408}. The methods have been based on machine learning algorithms \cite{pang-etal-2002-thumbs} trained with corpora that were labeled and curated by human experts \cite{pang-lee-2004-sentimental}. 

Later on, denser text representations based on word embeddings were invented \cite{NIPS2013_9aa42b31,pennington-etal-2014-glove} and larger sentiment analysis corpora like the one based on IMDb movie reviews\footnote{\url{https://ai.stanford.edu/~amaas/data/sentiment}} \cite{maas-etal-2011-learning} or other datasets of song lyrics \cite{Cano:2017:MSA:3059336.3059340,10.1145/3220228.3220229} were published. Better results were achieved using models based on neural networks \cite{kim-2014-convolutional,10.1007/978-3-319-77703-0_34}.

The recent developments based on pre-trained language models further improved the results \cite{hoang-etal-2019-aspect,10.1145/3503161.3548306}. These large models are pre-trained on huge amounts of unlabeled texts, but need to be fine-tuned with labeled corpora which are mostly in English language. Sentiment analysis research in low-resource languages is hindered by the lack of such corpora. In particular, no labeled and curated corpus for sentiment analysis of Albanian texts has been released yet.

\section{AlbMoRe Corpus} %

\begin{table}
\centering
\begin{tabular}{l c c}
\hline
& \textbf{Characters} & \textbf{Tokens}\\
\hline
Minimum & 28 & 5 \\
Maximum & 140 & 31 \\
Average & 75.5 & 15.2 \\\hline
\end{tabular}
\caption{Movie review length statistics.}
\label{tab:lenth}
\end{table}

User movie reviews are non-professional opinions that users post in social networks or websites about movies they watch. They are very popular today and come in the form of a numerical assessment, verbal (written) description or both. IMDb offers a large source of such reviews which have been used to create different research corpora.  

For building AlbMoRe, 67 movies listed in IMDb were chosen. These movies have been premiered in the late 80s, in the 90s and in the 00s. One criterion for choosing the movies was the maximization of genre diversity. To that end, the list includes movies of different genres such as \emph{action}, \emph{romance}, \emph{thriller}, \emph{fiction}, \emph{adventure}, \emph{comedy}, \emph{drama}, \emph{horror} and even a \emph{cartoon}.  

Another selection criterion was the length of the review text. Only reviews of at least one full and properly formatted sentence and at most four sentences long were collected. Since emotions of longer texts are usually harder to analyse, reviews of more than four sentences (which are numerous) were ignored. The review length statistics of the corpus are shown in Table~\ref{tab:lenth}. 

The user reviews of each movie were found in IMDb and initially ranked in descending and ascending order based on their star rating. Reviews of 10, 9 or 8 stars were considered as candidates for obtaining \emph{positive} samples. Similarly, reviews of 1, 2 or 3 stars were considered as candidates for obtaining \emph{negative} samples.   

The text descriptions of the candidate reviews were carefully read and translated in Albanian. At the same time, they were also labeled as either \emph{positive} or \emph{negative}. For each movie, an equal number of positive and negative reviews (from five to ten in each case) were collected, resulting in a fully balanced corpus with 400 \emph{positive} reviews and 400 \emph{negative} ones.

\begin{table}
\centering
\begin{tabular}{p{5.5cm} | c }
\hline
\textbf{Review} & \textbf{Polarity}\\
\hline
E adhuroj këtë film, edhe pasi e kam parë sa e sa herë. Skenat epike me aksion në beteja janë fantastike. & positive \\\hline
Aspak origjinal! Ka shumë gabime historike që e zbehin interesin për këtë film. & negative \\\hline
\end{tabular}
\caption{Illustration of two data samples.}
\label{tab:samples}
\end{table}

Table~\ref{tab:samples} illustrates two samples of AlbMoRe corpus which belong to the movie ``Gladiator'', premiered in the year 2000. The first review is positive and the second one is a negative.

\section{Preliminary Experimental Results} \label{sec:results}

This section presents the results of some basic experiments that were run using AlbMoRe corpus and a few traditional machine learning models for classification. Since these results are intended only as simple baselines for future studies, no advanced models were tried and no attempts for optimizations were made. 

\subsection{Preprocessing and Vectorization} \label{ssec:preproc}

Before feeding the text part of the AlbMoRe samples to the classification algorithms, a few pre-processing steps were performed. The texts were first tokenized, separating the words form the punctuation and special symbols. This operation results in loss of white-space symbols such as `\verb|\n|' or `\verb|\t|' which are actually not necessary. Furthermore, consecutive (two or more) spaces were replaced with a single space. All symbols were also lower-cased, which helps to reduce the vocabulary (set of unique words). No other text pre-processing steps like stemming or lemmatization were applied. Finally, Tf-Idf with default parameters was chosen for vectorizing the words.

\subsection{Classification Algorithms} \label{ssec:algorithms}

A few preliminary experiments were run on the corpus, trying four traditional machine learning algorithms. One of them is SVM (Support Vector Machine) which has been successfully used for both classification and regression tasks since the nineties when it was invented \cite{cortes1995support}. It utilizes the concept of hard and soft margins which are separation hyper-planes to optimally separate the samples of different classes from each other. The addition of the \emph{kernel} parameter enables SVM to perform well even on data that are not linearly separable by transforming the feature space \cite{Kocsor:2004:AKF:1008633.1008643}.
Logistic regression is another algorithm which despite being simple, yields good results on a high number of tasks. It makes use of the logistic function to determine the probability of samples pertaining to classes. It is also one of the fastest algorithms to train.

Decision trees have been around since many years and are based on a hierarchical tree structure with branches which represent the values of the analysed features and nodes which represent states or decisions \cite{quinlan:induction}. They usually work well when the data consist of different types of features mixed together.    

Finally, random forest is an Ensemble Learning method that was also invented in the 90s \cite{Ho:1995:RDF:844379.844681}. It computes the average of the results obtained from multiple decision trees, providing lower variance.  
\begin{table}
\centering
\begin{tabular}{l c}
\hline
\textbf{Model} & \textbf{Accuracy}\\
\hline
Support Vector Machine & 0.925 \\
Logistic Regression & 0.915 \\
Decision Trees & 0.81 \\
Random Forest & 0.875 \\\hline
\end{tabular}
\caption{Sentiment analysis results.}
\label{tab:results}
\end{table}

\subsection{Discussion} \label{ssec:discussion}

Each of the four supervised learning algorithms was trained with its default parameters on the 600 training samples of AlbMoRe and was tested on the respective 200 test samples. The accuracy scores obtained for each of them are shown in Table~\ref{tab:results}. As we can see, SVM leads with an accuracy of 92.5\,\%. Logistic Regression follows closely with an accuracy of 91.5\,\%. The two algorithms based on trees lag behind. Decision trees are especially weak reaching an accuracy of 81\,\% only. Random forest performs better, providing an accuracy of 87.5\,\%.

\section{Conclusions} \label{sec:conclusions}

Research on computational linguistics or natural language processing tasks such as sentiment analysis requires corpora which are not available for every language. To foster research on sentiment analysis of Albanian texts, this work creates and presents AlbMoRe, a corpus of movie reviews collected from IMDb. It consists of 800 text samples labeled as \emph{positive} or \emph{negative}. A set of experiments and the respective results is also presented. They should serve as baselines for future research.  

\bibliography{anthology,custom}
\bibliographystyle{acl_natbib}




\end{document}